\RequirePackage{xcolor}
\documentclass[journal,twoside]{ieeecolor}
\usepackage{tmi}
\let\labelindent\relax
\usepackage{cite}
\usepackage{algorithmic}
\usepackage{amsmath,amssymb,amsfonts}
\usepackage{algorithmic}
\usepackage{graphicx}
\usepackage{textcomp}
\usepackage{times}
\usepackage{epsfig}
\usepackage{graphicx}
\usepackage{amsmath}
\usepackage{amssymb}
\usepackage{graphicx}
\usepackage{amsmath}
\usepackage{amssymb}
\usepackage{booktabs}
\usepackage{url}

\usepackage{tikz}
\usepackage{comment}
\usepackage{amsmath,amssymb} % define this before the line numbering.
\usepackage{soul,color}
\usepackage{graphicx}
\usepackage{amsmath}
\usepackage{amssymb}
\usepackage{booktabs}
\usepackage{cuted}
\usepackage{capt-of}
\usepackage{comment}
\usepackage{breqn}
\usepackage{amsmath}
\usepackage{multirow}
\usepackage{enumitem}
\usepackage{makecell}
\usepackage{enumitem}
\begin{document}
\title{Disruptive Autoencoders: Leveraging Low-level features for 3D Medical Image Pre-training}
\author{Jeya Maria Jose Valanarasu\textsuperscript{1,3}~~~~~~Yucheng Tang$^{1}$~~~~~~Dong Yang$^{1}$~~~~~~Ziyue Xu$^{1}$~~~~~~Can Zhao$^{1}$\\
Wenqi Li$^{1}$~~~~~~Vishal M. Patel$^{3}$~~~~~~Bennett Landman$^{2}$ \\
Daguang Xu$^{1}$~~~~~~Yufan He$^\dagger$$^{1}$~~~~~~Vishwesh Nath$^\dagger$$^{1}$\\
$^1$NVIDIA~~~~~~$^2$Vanderbilt University~~~~~~$^3$Johns Hopkins University\\
\thanks{Parts of the work were done when Jeya Maria Jose Valanarasu was at NVIDIA for an internship. e-mail: jmjose@stanford.edu}
\thanks{$^\dagger$ co-senior advising}
% \thanks{e-mail: jmjose@stanford.edu}
}

\maketitle

\begin{abstract}
Harnessing the power of pre-training on large-scale datasets like ImageNet forms a fundamental building block for the progress of representation learning-driven solutions in computer vision. Medical images are inherently different from natural images as they are acquired in the form of many modalities (CT, MR, PET, Ultrasound etc.) and contain granulated information like tissue, lesion, organs etc. These characteristics of medical images require special attention towards learning features representative of local context. In this work, we focus on designing an effective pre-training framework for 3D radiology images. First, we propose a new masking strategy called local masking where the masking is performed across channel embeddings instead of tokens to improve the learning of local feature representations. We combine this with classical low-level perturbations like adding noise and downsampling to further enable low-level representation learning. To this end, we introduce  \textbf{Disruptive Autoencoders}, a pre-training framework that attempts to reconstruct the original image from disruptions created by a combination of local masking and low-level perturbations. Additionally, we also devise a cross-modal contrastive loss (CMCL) to accommodate the pre-training of multiple modalities in a single framework. We curate a large-scale dataset to enable pre-training of 3D medical radiology images (MRI and CT). The proposed pre-training framework is tested across multiple downstream tasks and achieves state-of-the-art performance. Notably, our proposed method tops the public test leaderboard of BTCV multi-organ segmentation challenge. 
\end{abstract}


\section{Introduction}
\begin{figure}[htbp]
    \centering
    \includegraphics[width=0.75\linewidth, page=2]{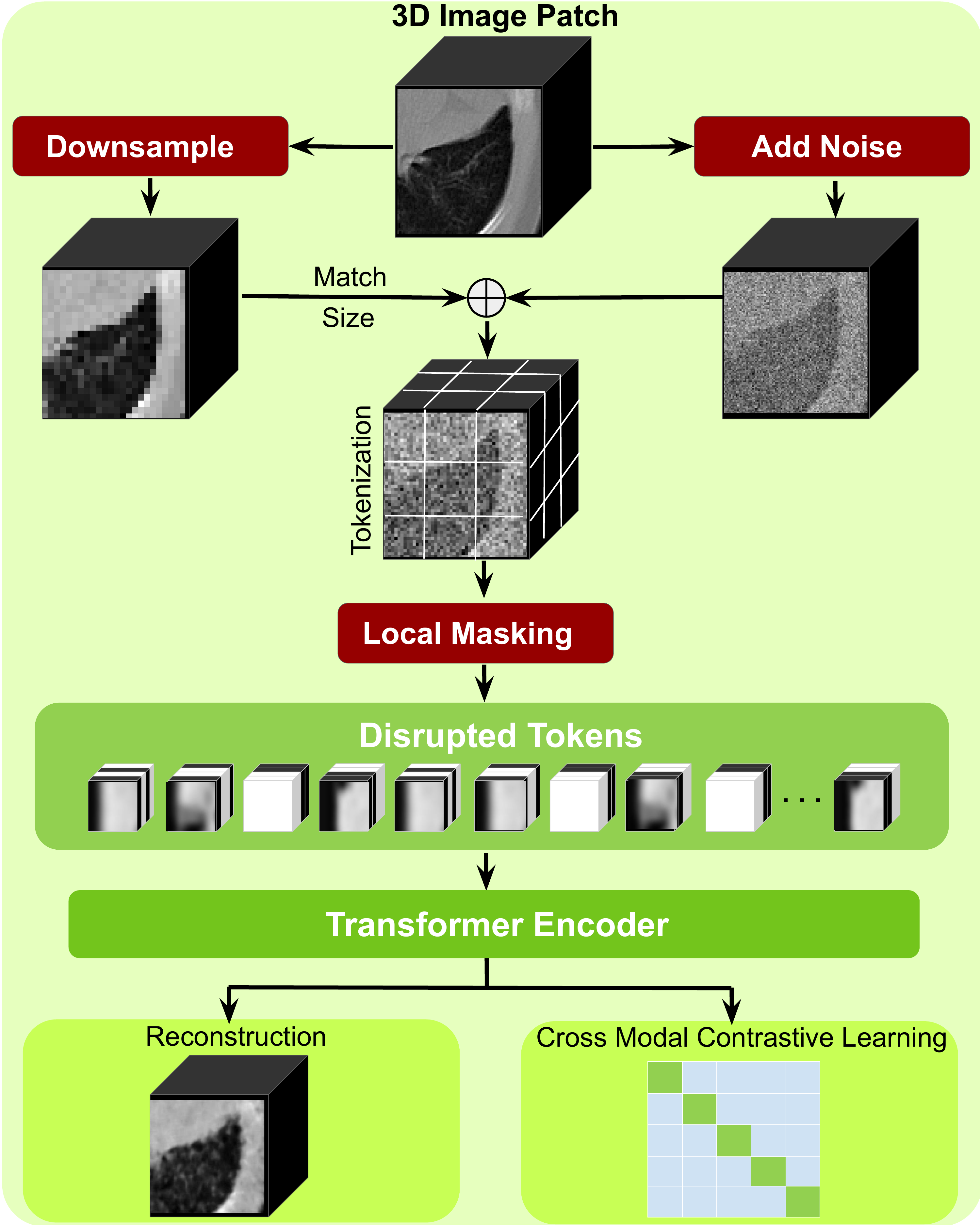}
    \caption{Disruptive Autoencoders: Here, we disrupt the 3D medical image with a combination of low-level perturbations - noise and downsampling, followed by tokenization and local masking. These disrupted tokens are then passed through a transformer encoder and convolutional decoder to learn to reconstruct the original image. Our method also includes cross modal contrastive learning to bring in modality-awareness to the pre-training framework. This can act as an effective pre-training strategy to extract meaningful low-level representations for 3D medical image analysis.}
    \label{fig:overview}
    \vspace{-0.5em}
\end{figure}

Inception of transformers \cite{vaswani2017attention, dosovitskiy2020image} has led to a significant shift from convolution neural network (ConvNet) based methods \cite{he2016deep,ronneberger2015u,krizhevsky2017imagenet} to transformer-based methods \cite{dosovitskiy2020image,liu2021swin, valanarasu2021medical, valan2022, chen2021transunet} for many computer vision applications. However, the fact that pre-training plays an irreplaceable role in model development has not changed in the past decade \cite{radford2021learning}.
Model weight initialization is an important step in training deep neural networks \cite{kumar2017weight} as good starting weights are necessary for efficient training towards a particular task. In the ConvNet era, pre-training on the ImageNet dataset \cite{deng2009imagenet} played a significant role in developing models for various downstream tasks. Most of the network architectures offer ImageNet weights as the starting weights for further training. In Vision Transformer (ViT) \cite{dosovitskiy2020image}, pre-training is proved to be even more fruitful due to its higher model capacity. It was observed that pre-training ViT on large-scale datasets like JFT300M \cite{sun2017revisiting} could provide superior performance than pre-training on ImageNet. Curation and collection of such large-scale datasets is thus pivotal to advancement of various computer vision sub-fields.

Pre-training for natural computer vision tasks is usually not constricted by the availability of data as natural images are abundant and there is no scarcity and less restrictions in obtaining them. Unfortunately, the same does not translate to medical images as they are scarce (acquisition cost is high) and also difficult to  obtain (requires specialized hardware). There is complexity involved to release them publicly due to heavy privacy regulations \cite{saliba2012telemedicine}. There have been recent efforts in creating a large-scale dataset for medical imaging \cite{ghesu2022self}, there still does not publicly exist a high-resolution dataset containing multiple modalities. It should be noted that while natural images are in general standardized RGB, medical images acquired by different modalities through different manufacturers' hardware with different acquisition parameters can differ significantly from each other. The characteristics of medical images are dependent on several factors, including modalities used for the specific diagnostic task, and the site where the image is acquired. Hence it is extremely challenging to build a generic pre-training framework for medical images that is capable of factoring in multiple possible variations simultaneously.

\begin{figure}[htbp]
    \centering
    \includegraphics[width=0.8\linewidth]{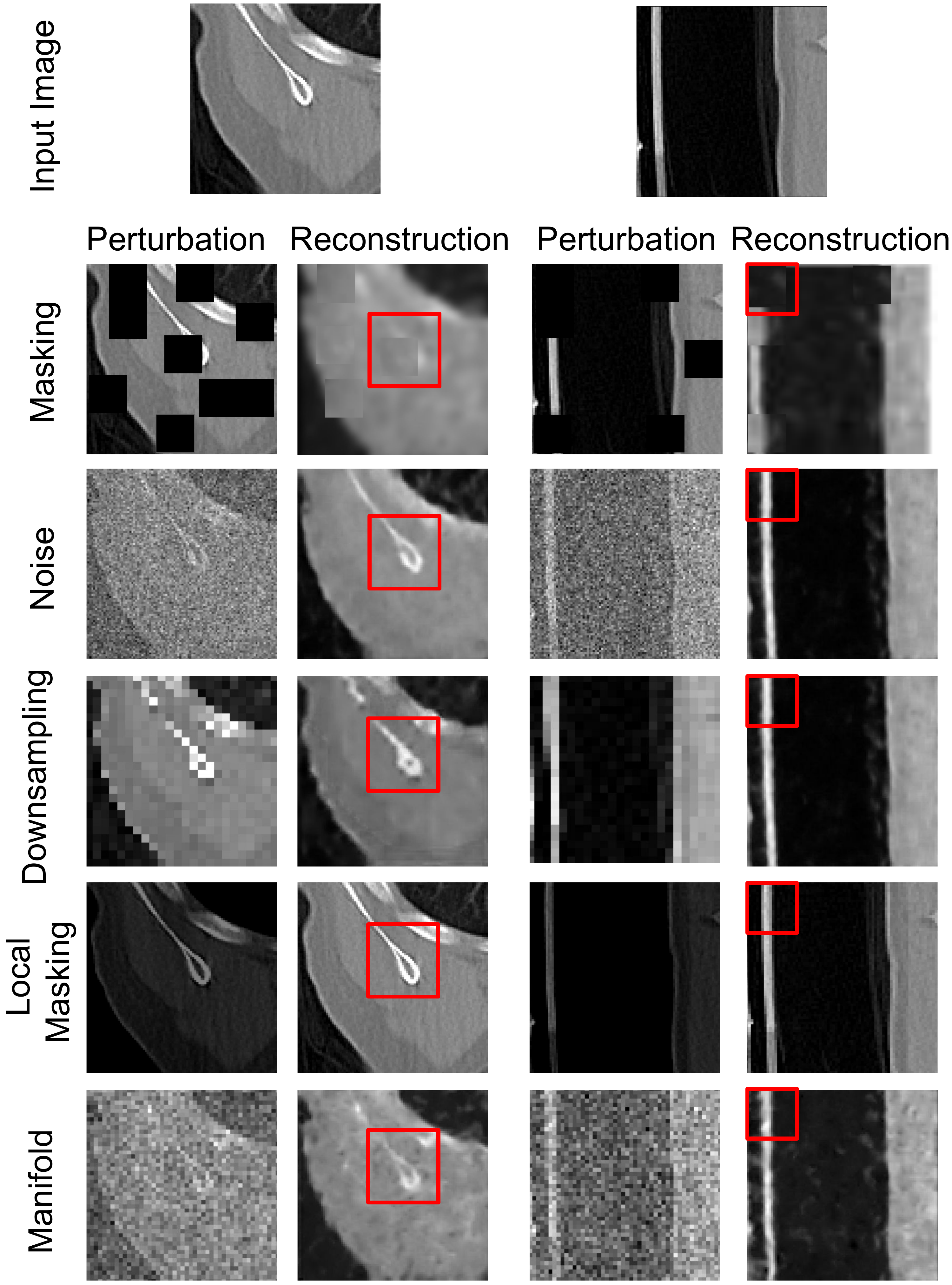}
    \caption{Comparison of reconstruction quality. It can be observed that masked image modelling produces a coarse reconstruction for radiology without local context while the proposed disruptions in this work obtain sharper reconstructions recovering meaningful fine details.}
    \label{fig:recons}
    \vspace{-1em}
\end{figure}

Recently,  masked image modelling methods \cite{he2022masked} have gained significant traction as an efficient self-supervised pre-training framework. They are used to develop robust pre-trained models that can generalize well to the downstream tasks. Masked Auto-Encoders MAEs learn a feature representation by trying to reconstruct the original image while masking out randomly selected tokens in the input space. MAEs are designed specifically for transformers as masked tokens help reduce the computation. However, improvements are needed in medical imaging domain: while trying to adopt vanilla MAEs for medical images, we observed that although MAEs do lead to a performance boost for further finetuning, the reconstructions were poor and most of the anatomical structures were missing after reconstruction. For example, in Fig. \ref{fig:recons} (left), it can be seen that MAEs cannot reconstruct the bones and other fine structures.  This has also been observed in some recent works \cite{hatamizadeh2022unetformer, zhou2022self}. Unlike natural images, most of the vital information in medical images are in the fine details (e.g. small lesions, finer boundaries of organs, tiny structures of bones that need to be delineated etc). To solve such tasks which require understanding fine details in an image, the focus is usually on extracting meaningful low-level features. In the context of features in different levels, computer vision tasks can be broadly classified into - low, mid, and high-level tasks depending on the type of information needed to solve the task. High level tasks like object recognition \cite{zhao2019object} and scene understanding \cite{silberman2012indoor} usually need a semantic understanding of the image. Low-level tasks like edge detection  and image matching  on the other hand, require a more fine understanding of the image. The type of features extracted by a deep network to solve low-level tasks are usually different from features extracted to solve high-level tasks \cite{nguyen2019understanding, valanarasu2020kiu}. Since MAEs can be considered as a form of inpainting \cite{xie2012image} which is regarded as a high-level task; most of the features extracted are mid to high-level features resulting in a coarse reconstruction as we observed in the outcome. This makes MAEs sub-optimal for pre-training medical images as features representative of crucial low-level information go missing.

In this work, we focus on designing an effective pipeline for pre-training on 3D medical volumes.  First, we design a new pre-training strategy that is better than MAEs at extracting low-level details. MAEs mask tokens at different places at a global level, making it difficult for the network to predict reconstructions while preserving local details. To this end, we introduce local masking where we do not mask at the token dimension but at the channel embedding dimension. Unlike MAEs, certain amount of masking is done to all tokens as only the channel embeddings are perturbed (visualized in Fig.~\ref{fig:overview} and \ref{fig:local_mask}). In this way, we mask out certain portion of information for all tokens throughout the input, helping the network reconstruct sharp details and learn better local context. We also explore using various low-level vision tasks like denoising and super-resolution for pre-training. We observe that these tasks help extract better low-level features and result in sharper reconstructions. In summary, we introduce Disruptive Autoencoders (DAE) where we first create a combination of these perturbations (local masking, downsampling, and adding noise) to disrupt the tokens (visualized in Fig.~\ref{fig:overview}). Then, an autoencoder is trained to reconstruct the original medical volume from these disrupted tokens. DAEs result in sharper reconstructions, and a better performance on downstream tasks.  We also devise a contrastive loss for our framework in such a way it can discriminate between the features extracted from different modalities. This cross modal contrastive loss (CMCL) pulls together features of the same modality and pushes apart features of different modalities. In this way, the pre-training strategy extracts diverse features from different modalities while maintaining an awareness of each modality. 

To evaluate our proposed method DAE we curate a pre-training dataset of radiology images. The dataset includes 5 modalities: Computed Tomography (CT) and 4 modalities of Magnetic Resonance Imaging (MRI) (T1, T2, FLAIR, T1ce). We use this collection of 10,050 individual 3D volumes for pre-training. All selected  datasets are public so that the pre-trained weights can be used as a better initialization as compared to random for future MR and CT radiology tasks. We conduct experiments on multiple segmentation baselines to validate the effectiveness of our proposed pre-training method.

In summary, the following are the major contributions of this work: 

\begin{itemize}[noitemsep]

    \item We propose Local Masking, a new masking strategy which helps learn useful features to reconstruct local details like small anatomical and functional structures.

    \item We introduce \textbf{Disruptive Autoencoders},  which aim to reconstruct the original volume from tokens disrupted from a combination of low-level perturbations such as local masking, downsampling, and adding noise.

    \item Our framework proposes using a single pre-trained weight for various downstream tasks even for different modalities.
    
    \item We curate a public pre-training dataset for CT and MRI radiology images with over 10,000 3D volumes and conduct extensive experiments on multiple segmentation datasets and show state-of-the-art performance. The pipeline achieves best performance on a public multi-organ segmentation challenge leaderboard.
    
\end{itemize}

\section{Related Works}

\noindent \textbf{Self-supervised Pre-training}: Self-supervised pre-training approaches can be broadly categorized into two types:  1) generative and 2) discriminative. Generative methods focus on mapping the input to a latent space to learn a representation and then decode it back to a new space. Masked Autoencoders (MAE) \cite{he2022masked} propose a way to mask out some tokens and make the network reconstruct the original image back thus helping the model learn useful representative features. It uses an asymmetric encoder-decoder design by having a small transformer decoder to reduce the computation burden. Beit \cite{bao2021beit} proposed a masked image modeling task to pretrain vision transformer while using two views of the input: image patches as well as a discrete visual token. SimMIM \cite{xie2022simmim} simultaneously performs representation
encoding and pretext prediction, due to which the decoder
design can be changed to be as simple as a single dense layer. Masked feature prediction \cite{wei2022masked} proposes a technique where instead of the original image, manual features like Histogram of Gradients (HOG) are extracted to learn the representation. Latent contextual regressors and alignment constraints have been proposed to map the predicted representations of masked patches to the targets. Masked Pre-training has not been applied only for images but also for point-clouds \cite{yu2022point}, videos \cite{girdhar2022omnimae},  and multi-spectral images \cite{cong2022satmae}.

Discriminative pre-training methods try to design a discriminative
loss to differentiate between the features extracted for different inputs. Typically ground truth in the form of annotations or labels is not used for pre-training, pretext tasks like solving jigsaw puzzles \cite{noroozi2016unsupervised} or predicting rotation \cite{gidaris2018unsupervised} are used to extract meaningful information. It is also worthy to note CLIP \cite{radford2021learning} uses a contrastive loss for multi-modal data to learn robust visual representations from image and text pairs. Contrastive methods have been shown to be useful in many other multi-modal contexts \cite{zhang2022pointclip}.

\noindent \textbf{Pre-training for Medical Images:} Model Genesis \cite{zhou2019models} proposes a unified self-supervised learning framework using the recurrent anatomy in medical images. Azizi et al. \cite{azizi2021big} perform a stage-wise pre-training on natural images followed by task specific medical images.  A Multi-Instance Contrastive Learning based approach is proposed to perform self-supervised pre-training. Several other methods \cite{kalapos2022self} also follow similar contrastive strategies for specific medical imaging tasks. In \cite{tang2022self}, a self-supervised framework using a combination of contrastive coding,  rotation prediction, and inpainting was proposed for pretraining on CT volumes. MAE-based pre-training methods have been quickly adopted for self-supervised pre-training on medical images \cite{ zhou2022self,chen2022masked}. These works show that masked image modelling methods provides better performance than previous contrastive methods. Unlike these works, we propose a new pre-training setup which efficiently pre-trains on multiple modalities contrastively while also learning all the low-level anatomical details using an autoencoder to have a better representative power.

\section{Proposed Method}

In this section, we first explain our motivation by exploring why MAEs are sub-optimal for pre-training on 3D medical images. Then, we describe the proposed pre-training strategy - DAE, followed by training details and loss functions.

\subsection{What do MAEs lack for medical images?}

MAEs have shown impressive results for vision based image pre-training. To this end, we first adapted MAEs to operate on 3D volumes for pre-training medical volumes. However, we observed that the reconstruction quality was low as the reconstructions lose the finer anatomical details. Such observations are also seen in other works that try to use vanilla MAEs for medical image pre-training as depicted in Fig.~3. of\cite{hatamizadeh2022unetformer} \& Fig.~2. of \cite{zhou2022self}. Although using these pre-trained weights do improve downstream tasks, we argue that there is significant potential for further improvement in learning better representations as compared to MAEs for medical images. Coarse reconstructions might be sufficient to understand a high-level semantic understanding which is useful for classifying natural images. For tasks like segmentation of medical images, we postulate that coarse features result in poor reconstructions and are not sufficient to enable efficient fine-tuning.  MAEs lack in the aspect of learning features that reflect a deeper understanding of the medical image as the tokens are masked globally and no special attention is given to learn the local details.

\subsection{Disruptive Autoencoders}

 Medical images (especially radiology images) are different from natural images as the information contained in them like anatomical details and lesions are mostly fine-grained details instead of coarse structures. To this purpose, the network architecture need guidance to learn features which are representative of low-level details to have more useful representations of the medical image. To tackle these challenges, we propose DAE, a pre-training strategy that focuses on learning strong low-level representative features of the medical image to reconstruct local context. Here, we first take a cubic patch from a 3D medical volume, perturb and tokenize it to get disrupted tokens in 3D. The disrupted tokens are then passed through a transformer encoder to learn a feature representation. The latent features are passed through a decoder and are learned to reconstruct the original image back. The tokens are disrupted using a combination of different low-level perturbations: i) Local Masking ii) Adding Noise and iii) Downsampling. Local Masking is a novel masking strategy proposed by this work. Denoising and super-resolution (recovering downsampled images) are classic low-level tasks, we are one of the first to explore them as pre-training tasks for medical imaging motivated by the tasks ability to affect low-level finer features of the images. In the next sections, we discuss these in greater detail.

\subsubsection{Local Masking}

Transformers were designed to function in a sequence-to-sequence manner where the image is first tokenized. Tokenization can either be done with just linear layers \cite{dosovitskiy2020image} or convolutional kernels \cite{liu2021swin}. The 3D input images are of dimension $(H,W,Ch)$ where $H$, $W$, and $Ch$ denote the height, width, and number of channels in the image respectively. After tokenization, the tokens are of the dimensions $(N,C)$ where $N$ represents the number of tokens and $C$ denotes the embedding dimension.  Masked image modelling methods like MAE and SimMIM follow a token masking approach where some tokens $X$ out of $N$ are set to zero and the network tries to reconstruct the original image back. The percentage of tokens masked here is a hyper-parameter. Token masking approach done in MAEs can be considered a global masking approach as the entire token chosen to be masked is set to zero. To be more specific, the entire $C$ dimension of the chosen $X$ tokens that are to be masked are set to zero.
 
\begin{figure}[htbp]
    \centering
    \includegraphics[width=1\linewidth]{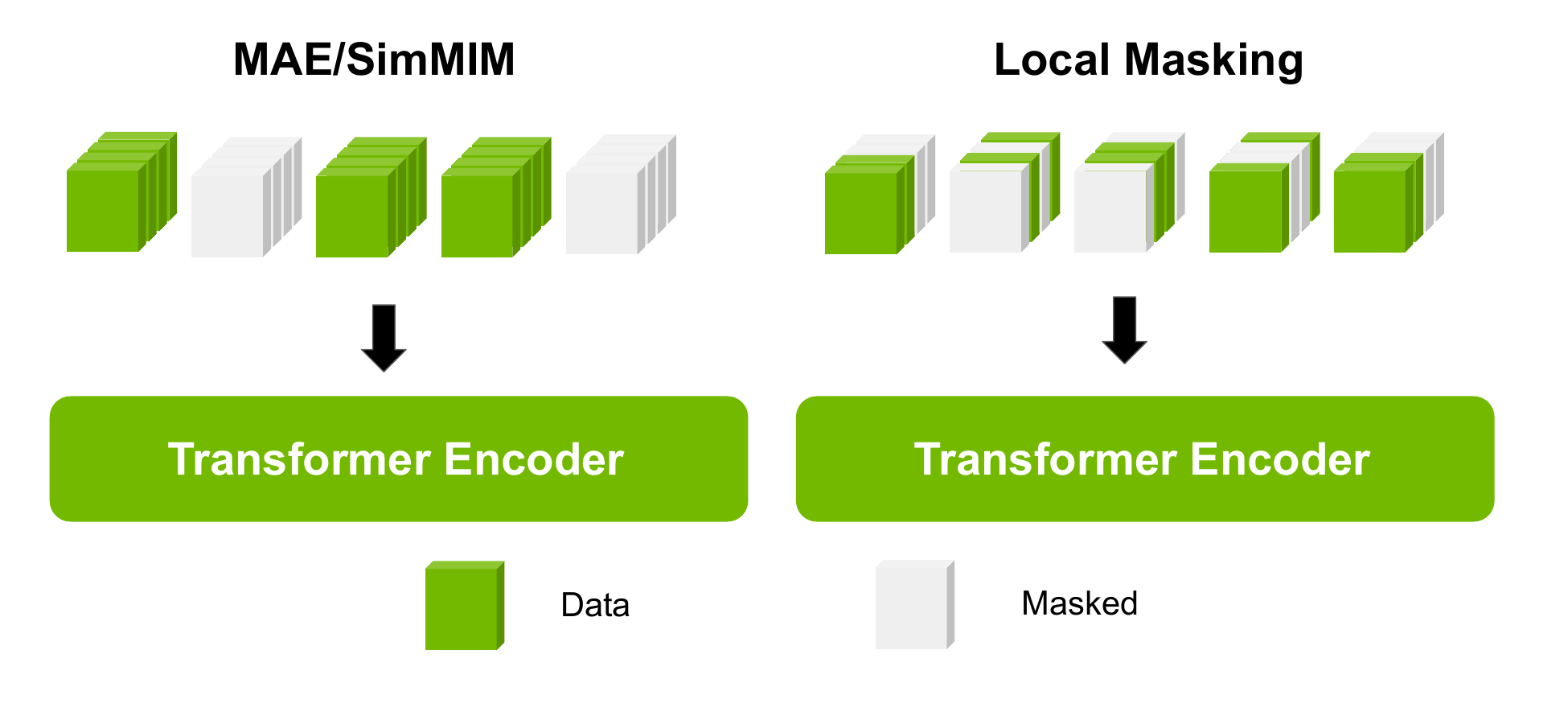}
    \vspace{-2em}
    \caption{Local Masking: Each token has channel embeddings, in MAE-based methods, the selected tokens are masked completely. In Local masking we only mask certain channel embeddings of each token instead of completely masking the token.}
    \label{fig:local_mask}
    \vspace{-1em}
\end{figure}

The entire $C$ dimension, when set to zero, disrupts the image globally, thereby directing the network to learn global context with the objective to reconstruct the original image. This globally disruptive process of setting $C$ does not always help in obtaining a good reconstruction for medical images as most of the information in medical images are not global but in the finer local details. Not being able to learn features representative of the local details like anatomy also affects the fine-tuning performance of MAEs. To this end, instead of masking $X$ tokens out of $N$, we propose to mask $X$ channels out of $C$ channel embedding dimension. This ensures that there is some information preserved for each token so that local details do not get completely destroyed. The perturbation is done locally as we set certain embeddings of each token as zero. We call this approach local masking as the masking is done to local details. Local masking has been visualized in Fig.~\ref{fig:local_mask}. The masking ratio $r$ here is a hyper-parameter which defines the percentage of $C$ embeddings being masked.

\subsubsection{Other Low-level Perturbations}

  While local masking helps us extract features representative of local context, we further try to employ other tasks like denoising and super-resolution for pre-training which would help learn more low-level information. Noise and low resolution are commonly found issues in realistic clinical medical acquisition pipelines and hence having them in the pre-training pipeline is meaningful \cite{zhao2019applications}.

\noindent \textbf{Adding Noise:}  %The noise present in the images is usually caused by various intrinsic or extrinsic conditions. 
Denoising is the task of restoring an original image from its noisy version. %The noise variation is a quite practical case for medical image acquisitions as depending upon the scanning hardware there are often varying levels of noise involved. 
To obtain a good denoised image, a model must be able to restore all local details of the image like edges and corners. To enable denoising as a pre-training task, we first add noise to the original input and try to restore the original image from the noisy input using the network. Given that the most common additive Gaussian noise, we define the perturbed input $\hat{x}$ as follows:
\setlength{\belowdisplayskip}{0pt} \setlength{\belowdisplayshortskip}{0pt}
\setlength{\abovedisplayskip}{0pt} \setlength{\abovedisplayshortskip}{0pt}
\begin{equation}
   \hat{x} = x + N(\mu,\sigma), 
\end{equation}
where $x$ is the original input and $N$ is the normal distribution. For each sample, we randomly sample from a normal distribution to get the noise. This way, there is no specific pattern for the network to easily restore the input. The mean and variance are hyper-parameters which can be used to control the noise level injected. 
% The correlation between the noise intensity during pre-training and downstream task performance has been studied in supplementary. %A low variance does not help learn any valuable information while a high variance makes it very difficult for the network to perform restoration. %{\bf{The correlation between the noise intensity during pre-training and downstream task performance has been studied in Section .}} 

\noindent \textbf{Downsampling:} Image Super-resolution refers to the task of enhancing the resolution of an image from a low-resolution (LR) to high-resolution (HR). Low-resolution images usually are blurred with sampling artifacts thus making it difficult to infer details from the image. Super-resolution is highly useful in medical imaging  as capturing high-resolution images for certain modalities like MR can be tricky due to long scanning time (as the resolution of the scan acquisition is increased, time and cost go higher exponentially). %Super resolution helps resolve this by generating high-resolution scans from otherwise low-resolution images.
To use super-resolution as a pre-training strategy, we first downsample the input image $x$ to get the LR image $\hat{x}$. We formulate this downsampling process 
as follows:
\begin{equation} 
    \hat{x} = D(x,\epsilon),
\end{equation}
where $D$ is the downsampling function and $\epsilon$ is the downsampling ratio. We use linear interpolation for downsampling. We follow a pre-upsampling super-resolution setup where we upsample the LR to same spatial dimension before converting it to HR. %The model is trained to recover the HR image $x$ from the LR input $\hat{x}$. Super-resolution is a low-level vision task as we need a features representing fine details to get a HR estimate from blurry  LR inputs. Thus, to produce a good HR image, the network needs to learn features that provide rich information to recover all the fine details of the image.

In DAEs, we use a combination of all the above perturbations.
We first add noise and downsample the image, then combine the two perturbations.
Then, the resultant 3D image is tokenized with a local masking strategy. We name these tokens as disrupted tokens and pass it to the transformer encoder. These features are then passed through a decoder for reconstruction and for cross-modal contrastive learning as explained in Fig.~\ref{fig:overview}. 

\subsection{Network Details and Training}

We use Swin-UNETR as our backbone architecture for all the experiments. Swin-UNETR follows a hierarchical vision transformer backbone that computes self-attention in an efficient shifted window partitioning scheme and has been proved to be suitable for multiple downstream medical imaging tasks \cite{tang2022self}. It has a U-shaped network design in which the extracted feature representations of the encoder are used in the decoder via skip connections at each
resolution. The encoder first creates a sequence of 3D tokens followed by transformer blocks consisting of window and shifted window self-attention mechanisms. This helps Swin-UNETR learn long-range dependencies in the image resulting in effective segmentation. For pre-training we use a combination of $\mathcal{L}_1$ reconstruction loss and the cross modal contrastive learning loss $\mathcal{L}_{CMCL}$.

First, we explain how to obtain $\mathcal{L}_{CMCL}$. Inspired from \cite{radford2021learning}  we train the network to
predict which of the $B \times B$ possible pairings
across a batch $B$. The goal is to maximize the cosine similarity of the embeddings of the true pairs
in the batch while minimizing the cosine similarity of the
embeddings of the incorrect pairings. We optimize a symmetric cross entropy loss over these similarity
scores. 
\begin{equation}
z_{sim} = z_i * z_i^T * exp(t),
\end{equation}
where $z_i$ is the mini-batch feature vector, $z_{sim}$ is the similarity matrix and $i$ is the mini-batch index. $t$ here is the temperature parameter and is set to 0.07. Note that the total number of data points is $N$ and each mini-batch has $B$ data points which means the mini-batch index $i$ goes from 1 to $N/B$. Now, we apply a simple binary cross entropy loss on the similarity matrix $z_{sim}$ to perform contrastive learning. We define CMCL loss as follows:
\begin{equation}
    \mathcal{L}_{CMCL} = \alpha *  CE(z_{sim}, z_{label}),
\end{equation}
where $CE$ represents the binary cross entropy loss, $\alpha$ represents the scale and $z$ denotes the latent feature vector.  CE loss is applied across each axis in the $z_{sim}$ matrix. $z_{label}$ is the label matrix that is created based on the positive pairs and negative pairs as per the meta-data. 

The total pre-training loss can be defined as:
\begin{equation}
    \mathcal{L}_{pretrain} =  \mathcal{L}_{1} +  \mathcal{L}_{CMCL}.
\end{equation}
The $\alpha$ in CMCL is set to 0.05 to match the range of both losses. We pre-train Swin-UNETR on a curated set of medical volumes without any labels then use those weights as starting weights for our downstream fine-tuning experiments. For finetuning experiments, we use a  Dice loss to train the model.

\section{Experiments}

\begin{table}[htbp]
\centering
\resizebox{0.27\textwidth}{!}{%
\begin{tabular}{c|ccc}
\hline
Method      & HD             & MSD            & Dice           \\ \hline
ResDSN      & 24.55          & 1.814          & 0.813          \\
3D FCN      & 38.59          & 4.601          & 0.792          \\
RandomPatch & 18.98          & 1.423          & 0.856          \\
nnUNET      & 18.39          & 1.335          & 0.888          \\
UNETR       & 39.05          & 1.275          & 0.891          \\
Swin-UNETR  & 20.53          & 0.810          & 0.918          \\
DAE (Ours)  & \textbf{16.82} & \textbf{0.654} & \textbf{0.921} \\ \hline
\end{tabular}
}
\caption{Leaderboard \protect\footnotemark
 results of the BTCV challenge on multi-organ segmentation. DAE achieves state-of-the-art performance in both free and standard competitions. Note: HD: Hausdorff Distance, MSD: Mean Surface Distance.}
 \label{leaderboard}
\end{table}

\footnotetext{\url{https://www.synapse.org/#!Synapse:syn3193805/wiki/217785}}

\subsection{Datasets}

\noindent \textbf{Pre-training Dataset:} We combine various public radiology CT and MRI datasets BraTS21 \cite{ bakas2018identifying}, LUNA16 \cite{setio2015automatic}, TCIA Covid19 \cite{desai2020chest}, HNSCC \cite{grossberg2018imaging}, TCIA Colon \cite{johnson2008accuracy}, and LiDC \cite{armato2011lung} to construct our
pre-training dataset. The corresponding number of 3D
volumes for brain, chest, abdomen and head/neck  volumes are $1,310 \times 4$ (4 modalities), 2,018, 1,520
and 1,223, respectively. The number of brain MRI volumes of each modalities T1, T2, T1ce and FLAIR is 1,310.

\begin{table}[htbp]
	\centering
		
		\resizebox{1\linewidth}{!}{%
	\begin{tabular}{c|c|c c c c c |c}
		\Xhline{2\arrayrulewidth}

	Data	&Method    & Fold1 &Fold2 &Fold3 &Fold4 &Fold5 &Avg     \\ \Xhline{2\arrayrulewidth}

			&Scratch            & 0.840 & 0.848    & 0.826    & 0.838 &0.844 & 0.839                  \\
		100 \%	&MAE            & 0.836  & 0.845    & 0.826     & 0.837 & 0.846 & 0.838                 \\
		&DAE           & 0.843 & 0.850    & 0.826     & 0.841 & 0.850 & \textbf{0.842}             \\\Xhline{2\arrayrulewidth}
			&Scratch           & 0.654 & 0.693    & 0.650    & 0.645 &0.667 & 0.662                 \\
		50 \%	&MAE            & 0.660  & 0.694    & 0.658     & 0.640 & 0.669 & 0.664                 \\ 
		&DAE           & 0.680 & 0.713    & 0.663     & 0.641 & 0.670 & \textbf{0.673}             
		\\\Xhline{2\arrayrulewidth}
			&Scratch            & 0.610 & 0.620    & 0.619    & 0.598 &0.644 & 0.618                 \\
		20 \%	&MAE            & 0.615  & 0.640    & 0.610     & 0.616 & 0.642 & 0.625                 \\
		&DAE           & 0.636 & 0.671    & 0.648     & 0.633 & 0.669 & \textbf{0.651}             \\
		\Xhline{2\arrayrulewidth}

	\end{tabular}
	}
 
	\caption{Dice score on 5-fold cross validation on  FeTA dataset.}

	\label{feta}
  \vspace{-1em}
\end{table}

\noindent \textbf{Finetuning Datasets:} i) \textbf{BTCV:}  Beyond the Cranial Vault (BTCV) abdomen
challenge dataset \cite{landman2015miccai} consists of abdominal CT scans of 30 subjects. The annotations contain 13 organs which are annotated by interpreters under
supervision of radiologists at Vanderbilt University Medical
Center. Each CT scan is acquired with contrast enhancement
phase at portal venous consists of 80 to 225 slices with
$512 \times 512$ pixels and slice thickness ranging from 1 to 6 $mm$.
The multi-organ segmentation problem is formulated as a 13-class segmentation task. 

\begin{figure*}[htbp]
    \centering
    
    \includegraphics[width=1\linewidth]{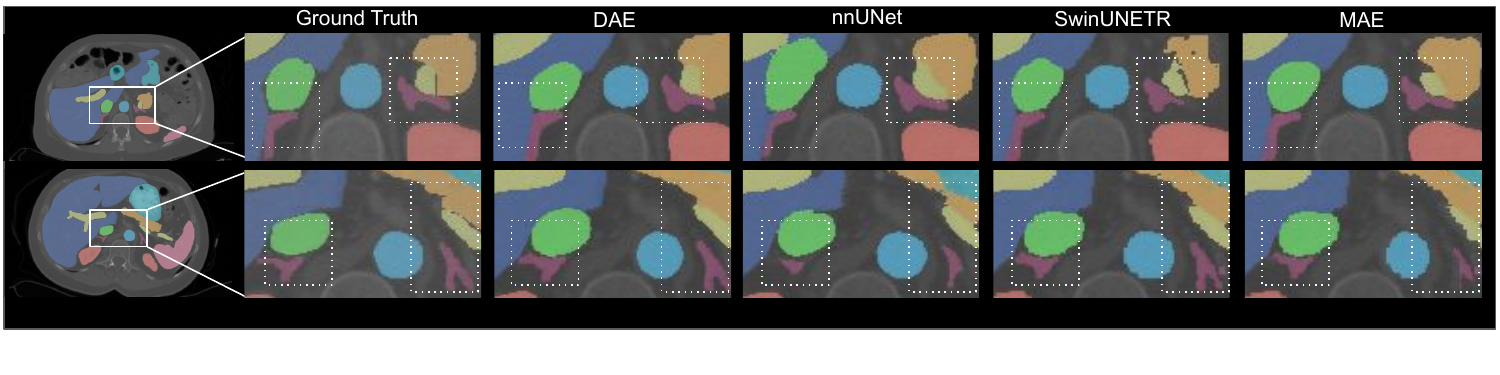}
  \vspace{-2em}
    \caption{Qualitative visualizations of the proposed DAE and baseline methods for BTCV dataset on two  randomly chosen subjects.}
    
    \label{fig:btcv_qual}
    \vspace{-1em}
\end{figure*}

\noindent ii) \textbf{FeTA}: Fetal Tissue Annotations dataset (FeTA) \cite{payette2021automatic} consists of publicly available database of 120 manually segmented pathological and neurotypical fetal MRI T2-weighted brain volumes. These  volumes are across a range of gestational ages (20 to 33 weeks) and are segmented into 7 different tissue
categories (external cerebrospinal fluid, grey matter, white matter, ventricles, cerebellum, deep grey matter, brainstem/spinal cord). The training images were acquired at two different sites with 1.5T and 3T MR scanners. The data were provided with histogram based matching and zero-padded for $256 \times 256 \times 256$ voxels. Data of both sites was also sampled at 0.5 $mm$ isotropic spacing as per challenge design. The dataset was split into 5-folds of 80/20 for training and validation.%, more details can be referred at the challenge website\footnote{https://feta.grand-challenge.org/Data/}.

\subsection{Implementation Details}

Our deep learning models were implemented in PyTorch \cite{NEURIPS2019_9015} and MONAI. 
For pre-training experiments, we used a batch-size of 2 per GPU. The volumes were randomly cropped into $96 \times 96 \times 96$ cubes while pre-training. We used an initial learning rate of $4e^{-4}$,
momentum of 0.9 and decay of $1e^{-5}$ for 20K iterations.  We trained the model using an AdamW \cite{loshchilov2017decoupled}
optimizer with a warm-up cosine scheduler of 500 iterations. We use hyper-parameters $r=60 \%$, $\sigma = 0.1$, $\epsilon=4$ when training the DAE.

For BTCV fine-tuning experiments, a five-fold cross validation strategy is used to train the models. The models were trained for 600 epochs with a learning rate of $4e^{-4}$ and the batch-size was set to 4 per GPU. Multiple augmentations such as gaussian noise, contrast scaling, zoom and random flipping across the axis were utilized. We select the best model in each fold and ensemble their outputs for final segmentation predictions.  For  FeTA, the intensities were normalized to a scale of 0 to 1. The learning rate was set to $4e^{-4}$ and batch size was set to 4 per GPU. All models were trained for 600 epochs, which was determined by convergence for the full dataset. Augmentations like random flipping on all 3 axes, Gaussian noise etc. were utilized during the training process. The final layer of the network is also changed from the pre-training configuration to accommodate the fine-tuning task at hand. For  FeTA, the output channels were set to 8 (including background) as per the dataset.  All pre-training and fine-tuning models are trained using NVIDIA DGX-1 V100 servers with 8 and 4 GPUs, respectively.

\subsection{Comparison with Previous Works}

% \footnotetext{https://www.synapse.org/#!Synapse:syn3193805/wiki/217785}

\begin{table*}[htbp]
		 \begin{minipage}{0.33\linewidth}
		 \centering
		 \vspace{0em}
		\resizebox{0.83\linewidth}{!}{%
	\begin{tabular}{c|c}
		\Xhline{2\arrayrulewidth}
		Pre-training Method           & Dice ($\uparrow$)                           \\ \Xhline{2\arrayrulewidth}
    Scratch &  0.8343    \\
		Contrastive Coding \cite{chen2020simple} &  0.8367    \\
		Rotation Prediction \cite{tang2022self}            & 0.8356      \\
		MAE \cite{he2022masked}         & 0.8448               \\
		DAE (Ours)       & \textbf{0.8512}                \\
		\Xhline{2\arrayrulewidth}
	\end{tabular}
	}
	\caption{Comparison with previous methods.}
	\label{comp}
	\end{minipage}%
		\begin{minipage}{0.33\linewidth}
		\centering
		\resizebox{1\linewidth}{!}{%
	\begin{tabular}{c|c}
		\Xhline{2\arrayrulewidth}
		Pre-training  Disruption Strategy           & Dice ($\uparrow$)                           \\ \Xhline{2\arrayrulewidth}
    Scratch &  0.764    \\
		Noise &  0.807    \\
		Downsampling             & 0.803      \\
		Local Masking          & 0.793               \\
		DAE         & \textbf{0.831}                \\
		\Xhline{2\arrayrulewidth}

	\end{tabular}
	}
	\caption{Ablation Study on DAE.}
	\label{ablation}
\end{minipage}%
\begin{minipage}{0.33\linewidth}
\centering
\vspace{0.5em}
		\resizebox{0.7\linewidth}{!}{%
	\begin{tabular}{c|c}
		\Xhline{2\arrayrulewidth}

		Pre-training \\ Strategy           & Dice ($\uparrow$)                           \\ \Xhline{2\arrayrulewidth}
    Scratch &  0.789    \\
		DAE w/o CMCL &  0.841    \\
		DAE with CMCL             & \textbf{0.849}      \\
		
	\Xhline{2\arrayrulewidth}
	\end{tabular}
	}
	\vspace{0.8em}
	\caption{Impact of CMCL.}
	\label{cmcl}
 \end{minipage}
 % \vspace{-1em}
\end{table*}
We compare our proposed method with previous self-supervised methods like contrastive coding \cite{chen2020simple, tang2022self}, rotation prediction \cite{taleb20203d, tang2022self}, and masked image modelling methods \cite{he2022masked, xie2022simmim}. We use Swin-UNETR as our network backbone for all these experiments. We note that MAE and SimMIM are very similar to each other, the only difference being MAEs discard masked tokens while SimMIM includes them. So, we just utilize a masked image modelling configuration but with Swin-UNETR as the backbone and call this configuration MAE in all upcoming discussions.  For BTCV, we directly validated our predictions in the public leaderboard and so that we can compare our method with all previous backbone methods. For the leaderboard submissions, we submit to the free competition (no specific registration process required). We train all our models with 80 subjects (20\% as validation set), and evaluates on the 20 images test set with spacing resolution of $1 \times 1 \times 1$mm. Within the 80 images, 30 scans are from the public challenge data and 50 extra CT scans annotated by radiologists are used to boost the training performance. We perform 4 rounds of five-fold cross validation experiments and ensemble models to obtain the final prediction. The ensemble process are effective to exclude outliers. In addition, test time augmentation, boundary smoothing and connected component analysis are used for post-processing the labels. Note that this pipe-line for BTCV leaderboard submission is similar to the previous approaches like \cite{tang2022self} for fair comparison.

These results are tabulated in Table \ref{leaderboard}. We note that our proposed method performs the best and outperforms all the previous baselines. Specifically, we note that we outperform Swin-UNETR \cite{tang2022self} which also uses SSL pre-training consisting of 3 different SSL pretext tasks.  In particular, we obtain a significant improvement in terms of Hausdorff Distance (HD) and Mean Surface Distance (MSD) compared to previous methods. We also conduct a paired t-test
between our BTCV test Dice scores and Swin-UNETR’s results. We obtain a two-tailed p-value of 0.0318 which shows
our improvement is statistically significant (p $\le$ 0.05). We also note we obtain better results than even recent methods like UniMiSS \cite{xie2022unimiss} which works on the same dataset.

We also present our results on five-fold cross validation of BTCV according to \cite{tang2022self} and compare our method with previous SSL methods in Table \ref{comp}. It can be observed that our method performs better than all the previous SSL baselines compared including MAEs. Note that even the combination of the 3 pretext SSL tasks proposed in Swin-UNETR \cite{tang2021high} achieves 0.8472 dice score which is less than our obtained result. For  FeTA , we perform five-fold cross validation experiments and compare our method with from scratch and MAE. These results are tabulated in Tables \ref{feta}. We also show additional experiments on 50$\%$, and 20$\%$ training data in finetuning to show the usefulness of our method. It can be observed that our method performs even better under low-data regime which shows the benefits of pre-training. We also visualize some sample qualitative results of the BTCV dataset in Fig.~\ref{fig:btcv_qual}. It can be observed that DAEs result in better segmentation predictions specifically performing better at segmenting small anatomy when compared to MAEs and other baselines.

\section{Discussion}

\noindent \textbf{Ablation Study:} In Table \ref{ablation}, we conduct an ablation study on DAE. We compare with each of the disruptions separately and then compare against a combination of them. These experiments are conducted on the first fold of PROMISE12 test set. For this task, we observe that all three methods perform better than scratch while local masking performs the best out of the three. The combination of the disruptions obtains better performance. We note that this trend totally depends on the downstream task but with a combination of these perturbations, the pre-trained weights are always better at performance than random initialization.

\noindent \textbf{Impact of CMCL:} To understand the impact of CMCL, we conduct an experiment on DAE with and without CMCL loss. This experiment is conducted on single fold of the BTCV dataset. It can be observed in Table \ref{cmcl} that CMCL provides a benefit in terms of fine-tuning performance.

\noindent \textbf{Empirical Analysis to prove low-level features matter:} Since  a major premise of this work is to improve pre-training pipeline by extracting better low-level features, we conduct a simple experiment to show that low-level features are the most important for fine-tuning. We use CKA~\cite{kornblith2019similarity} as the  feature similarity metric. CKA is used to represent the correlation between any two feature vectors in the latent space. For this experiment, we use MAE and DAE pretrained weights and finetune it on a single fold of BTCV. Now, we feed forward the test images to both the pre-trained model and the fine-tuned model. The CKA calculated between the features extracted from pre-trained model and fine-tuned model across different stages is reported in Table \ref{cka}. This value gives us an estimate of how much features changed from the initial pre-trained weights to the final fine-tuned model. It can be observed that the deeper layers have the least CKA meaning most of the high-level features have changed after fine-tuning. On the other hand, the early layers have a higher CKA meaning more low-level feature representations were retained from the pre-trained weights to the fine-tuned model. As high-level features undergo a relatively heavier change compared to low-level features after fine-tuning, therefore it is important to focus on learning stronger low-level features during pre-training.

\begin{table}[htbp]
	\centering
		
		\resizebox{0.7\linewidth}{!}{%
	\begin{tabular}{c |c c c c c}
		\Xhline{2\arrayrulewidth}

		Stage    & 1 &2 &3 &4 &5     \\ \Xhline{2\arrayrulewidth}

			MAE            & 0.068 & 0.013    & 0.012    & 0.006 & 0.001 \\
			
			DAE            & 0.091 & 0.012    & 0.011    & 0.003 & 0.001 

	\end{tabular}
	}
	\caption{CKA values across different levels of the network calculated between pre-trained and fine-tuned model.}
	
	\label{cka}
\end{table}

\noindent \textbf{Limitations:}  There is scope for improvement in picking the best combination of $\sigma, \epsilon, r$ for the pre-trained weights. A grid search on these parameters would be time-consuming but would lead to better pre-trained weights. Moreover, it can be noted that it takes a lot of time and compute to conduct pre-training experiments with large-scale datasets on dense medical volumes. Additionally, if the size of the pre-training dataset is increased, it is likely to obtain better weights.

\section{Conclusion}
In this work, we proposed a new pre-training framework for 3D medical images called Disruptive Autoencoders, where tokens are disrupted using a combination of perturbations: local masking, additive noise, and downsampling. In particular, local masking is a new masking strategy where the masking is performed across channel embeddings instead of tokens to improve the learning of local feature representations. DAE as a pre-training framework performs better than other pre-training strategies across multiple 3D segmentation datasets. 
% Notably, using our proposed method we also achieve the best performance in a public multi-organ segmentation challenge leaderboard and state-of-the-art results on the other public datasets.

%%%%%%%%% REFERENCES
\bibliography{tmi_arxiv}
\bibliographystyle{ieeetr}

\end{document}